%% file: main.tex
\pdfoutput=1

\documentclass[final,3p,times,twocolumn]{elsarticle}

\usepackage{lineno,hyperref}
\modulolinenumbers[5]

\usepackage{algorithm} 
\usepackage{algorithmic}  
\usepackage[algo2e]{algorithm2e}
\usepackage{booktabs}
\usepackage{amsmath}
\usepackage{subfig}
\journal{Journal of Knowledge-Based Systems}

\newcommand\epsliongreedy{\operatorname{\epsilon-greedy}}
\newcommand\zeromatrix{\operatorname{0-matrix}}
\newcommand\zerovector{\operatorname{0-vector}}
\usepackage{xcolor, soul}

\begin{document}
\begin{frontmatter}






\title{A clustering-based reinforcement learning approach for tailored personalization of e-Health interventions\footnote{}}

\author[mymainaddress,mysecondaryaddress]{Ali el Hassouni\corref{mycorrespondingauthor}}
\ead{a.el.hassouni@vu.nl}
\author[mymainaddress]{Mark Hoogendoorn}
\ead{m.hoogendoorn@vu.nl}
\author[myfourthaddress,mythirdaddress]{Martijn van Otterlo}
\ead{mail@martijnvanotterlo.nl}
\author[mymainaddress]{A. E. Eiben}
\ead{a.e.eiben@vu.nl}
\author[mysecondaryaddress]{Vesa Muhonen}
\ead{vmuhonen@mobiquityinc.com}
\author[myfifthaddress]{Eduardo Barbaro}
\ead{eduardo.barbaro@ibm.com}

\address[mymainaddress]{Vrije Universiteit Amsterdam, Department of Computer Science, Amsterdam, The Netherlands}
\address[myfourthaddress]{Radboud University, Department of Computer Science, Nijmegen, The Netherlands}
\address[mythirdaddress]{Open University, Department of Computer Science, Heerlen, The Netherlands}
\address[mysecondaryaddress]{Mobiquity Inc, Data Science and Analytics, Amsterdam, The Netherlands}
\address[myfifthaddress]{IBM, Cognitive and Analytics Benelux, Amsterdam, The Netherlands}

\begin{abstract}
Personalization is very powerful in improving the effectiveness of health interventions. Reinforcement learning (RL) algorithms are suitable for learning these tailored interventions from sequential data collected about individuals. However, learning can be very fragile. The time to learn intervention policies is limited as disengagement from the user can occur quickly. Also, in e-Health intervention timing can be crucial before the optimal window passes. We present an approach that learns tailored personalization policies for groups of users by combining RL and clustering. The benefits are two-fold: speeding up the learning to prevent disengagement while maintaining a high level of personalization. Our clustering approach utilizes dynamic time warping to compare user trajectories consisting of states and rewards. We apply online and batch RL to learn policies over clusters of individuals and introduce our self-developed and publicly available simulator for e-Health interventions to evaluate our approach. We compare our methods with an e-Health intervention benchmark. We demonstrate that batch learning outperforms online learning for our setting. Furthermore, our proposed clustering approach for RL finds near-optimal clusterings which lead to significantly better policies in terms of cumulative reward compared to learning a policy per individual or learning one non-personalized policy across all individuals. Our findings also indicate that the learned policies accurately learn to send interventions at the right moments and that the users workout more and at the right times of the day.

\bigbreak
\noindent $^{1}$ This paper is a significantly extended version of the work by el Hassouni et al., 2018 \cite{el2018personalization}. In this work, we include a related work section with the most recent related publications. We also describe our proposed methods in more detail. Additionally, we select and implement a different e-Health benchmark based on the HeartSteps data set for the evaluation of our methods. Furthermore, we perform more extensive experiments and analyses of the results and describe our simulation environment for e-Health in more detail. We estimate that around 75\% of the content is new. 
\end{abstract}

\end{frontmatter}

\input{introduction.tex}
\input{related_work.tex}
\input{approach}
\input{evaluation_environments}
\input{experimental_setup}
\input{results}

\input{discussion}
\input{Acknowledgment}
\bibliography{mybibfile}

\end{document}

%% file: introduction.tex
\section{Introduction}
\label{intro}

The amount of data being collected about people's health state and behaviour has seen a huge increase in the last decade \cite{muller2012health,andreu2015big,murdoch2013inevitable,herland2014review,bates2014big}. This information originates from many different sources ranging from medical devices and medical doctors at hospitals to smartphones, smartwatches, and other sensory devices people carry and use daily. Consequently, these devices are a good source of useful data and at the same time, they can be used to provide interventions to users directly \cite{raghupathi2014big}. In healthcare in general, and e-Health specifically, learning which interventions work best in varying situations is a very relevant and important problem. Generally, one-size-fits-all solutions, where different users may be provided with the same intervention, are shown to be less effective compared to approaches that rely on penalization where tailoring interventions towards (groups of) users is common (see e.g. ~\cite{kranzler2012personalized,schmidt2006does,simon2000randomised,curry1995randomized,chawla2013bringing}). The data collected around these users is being used to perform such personalization \cite{chawla2013bringing}.

\bigbreak
\bigbreak

Personalization \cite{fan2006personalization} of interventions poses several challenges. Firstly, the success of interventions is not immediately clear, and an emphasis should be placed on interventions that lead to a sustained improvement in the health state rather than quick wins \cite{byng2005using}. Secondly, interventions are typically composed of sequences of actions (e.g. multiple support messages or exercises) that should act in harmony \cite{greenberg2009adherence}. To address these challenges,  reinforcement learning (RL) (see e.g.~\cite{wiering2012reinforcement}) arises as a very natural choice (cf.~\cite{hoogendoorn2017machine}).

While the RL paradigm fits this setting very well, certain properties of RL do not. The algorithms typically require a substantial learning period before a suitable policy (specifying which intervention action to select in what situation) is found \cite{dulacarnold2019challenges}. In health settings in general, we do not have a sufficiently long learning period per user, and trying a lot of unsuitable actions can disengage users \cite{kreyenbuhl2009disengagement}. Hence, there is a need to substantially shorten the learning period. To establish this, we can either: $(1)$ start with an existing model (transfer learning, see e.g.~\cite{taylor2009transfer}) or $(2)$ pool data from multiple users that are similar to learn policies (cf.~\cite{zhu2017group}). While both are viable options, the latter one has not been explored for more complex and realistic health settings yet \cite{denHengstEtAlReviewPaper}.

Several avenues have been explored to shorten the learning period. Transfer learning (see e.g.~\cite{taylor2009transfer}) is one of them, where one learns a policy for one user (or across all users) which can be reused (and tailored) to other users. Recently, an RL algorithm that learns a policy for clusters of users has been proposed (cf.~\cite{zhu2017group}). In experiments, both approaches have shown to be viable to improve the learning speed.

In this paper, we extend our earlier work \cite{el2018personalization} that presents a cluster-based RL algorithm and evaluates it for a complex e-Health setting using a dedicated simulator we have built where interventions are sent to users to maximize a certain goal (e.g. working out). In this work, we perform more extensive experiments and analyses of the obtained results. Furthermore, we test the applicability of our methods in different e-Health scenario's and compare them to an e-Health setting from \cite{zhu2017group} that is based on the real-word HeartSteps \cite{dempsey2015randomised} dataset. We use $k$-Medoids clustering \cite{kaufman1987clustering} with Dynamic Time Warping (DTW) ~\cite{Berndt:1994:UDT:3000850.3000887} as the distance function to find suitable clusters, thereby automatically selecting a value for $k$ using the silhouette score \cite{ROUSSEEUW198753}. We learn policies over the clusters using both an online RL algorithm (Q-learning, cf.~\cite{watkins1992q}) and a batch algorithm (Least-Squares Policy Iteration (LSPI). cf.~\cite{lagoudakis2003least}). We compare the cluster-based approach to learning a single policy across all users and learning completely individualized policies. The aforementioned simulation environment we developed generates realistic user data for an e-Health setting. Here, the aim is to coach users towards a more active lifestyle. The simulator is made publicly available to allow for benchmarking and make it easier for others to evaluate novel RL approaches for this setting \footnote{An RL multi-agent simulation environment for e-Health \cite{ali_el_hassouni_2020_3826055}: www.github.com/alielhassouni/rl-multi-agent-simulation-for-e-health}.


In comparison with \cite{zhu2017group}, our approach relies on a more sophisticated and complex simulation environment where several types of users defined by a lifestyle schedule and personality are simulated with each their own behavioral profile and personal preferences which allows for highly personalized policies. Furthermore, we perform clustering using a state-of-the-art distance metric to learn optimal policies for clusters of users. We subsequently argue and empirically demonstrate that the stochasticity in the behavior of users makes the simulation environment a robust testbed for RL algorithms. 

This paper is organized as follows. We discuss related work Section 2 and present our cluster-based RL algorithm in Section 3. We continue with a description of the simulator we have developed in Section 4 along with a description of the HeartSteps benchmark. We then explain our experimental setup and our results in Sections 6 and 7 respectively. We close with a discussion.

%% file: related_work.tex
\section{Related work}

In recent years personalization using RL has seen a significant upward trend in many domains and especially in healthcare applications \cite{denHengstEtAlReviewPaper, ginsburg2001personalized, aspinall2007realizing, Zhao_2011, zamborlini2015inferring, simon2010personalized, shortreed2011informing, paper3_end_to_end_RL}. Judging from the systematic literature review by [den Hengst et al., 2020], several interesting statistics were found with regards to the application of RL for personalization problems \cite{denHengstEtAlReviewPaper}. For the papers that rely on RL for personalization, all information to base the personalization on was found to be accessible directly from data generated by the users of the RL system \cite{de2015line,tsiakas2015multimodal,baniya2018adaptive,bouneffouf2012hybrid}. Interestingly, only a small part of the papers considered the privacy and safety aspects of the application of RL for personalization \cite{goldberg2012q,de2015line,theocharous2015personalized}. As for the suitability of system behaviour towards users, in most cases this was derived from data instead of explicitly asking the users \cite{tsiakas2015multimodal,de2015line,theocharous2015personalized,baniya2018adaptive,el2019end,bragg2016optimal}. Furthermore, it is observed that a large percentage of publications across all domains rely on simulations for both policy development and algorithm evaluation \cite{denHengstEtAlReviewPaper}. Finally,  most RL applications for personalization develop one policy across all users and most of the remaining work develops one policy per user \cite{denHengstEtAlReviewPaper}.

We model the personalization system as an RL system that can act by sending \emph{interventions} to \emph{users}. The goal is to find a (stochastic) mapping from user states to interventions, by exploring possible strategies to do so, based only on evaluative feedback on performance. This use of RL for intervention strategies in health, wellness, coaching, and fitness applications is a relatively new development, although much other work has considered various \emph{nudging} approaches to stimulate human users to do particular things in various ways \cite{marteau2011judging, vlaev2016theory, sugden2009nudging, bucher2016nudging, hansen2016making}. To illustrate, \emph{adaptive persuasive systems} \cite{kaptein2013adaptive} have been tested in field trials, for instance to increase the effectiveness of e-mail reminders.

RL techniques \cite{wiering2012reinforcement,suttonbarto2nd} are ideally suited for sequential decision making problems in health interventions, \emph{dynamic treatment regimes} \cite{chakraborty2014dynamic}, or in motivational strategies in citizen science \cite{segal:citizen2018}. Work in this area has just begun to explore computational approaches. Several problems in (mobile) healthcare generate new challenges for RL, such as the problem of missing data, privacy, and especially the difficulty of interactive simulations with real human data \cite{dulacarnold2019challenges,denHengstEtAlReviewPaper}. For that reason we implemented a realistic simulator as an alternative data gathering option. A challenge is, however, to keep as close as possible to actual human data.

[Hochberg et al., 2016] compare RL -- in particular contextual bandits --  with static reminder policies to encourage diabetes patients through SMS interventions \cite{hochberg2016reinforcement}. [Raghu et al., 2017] combine continuous state-space models and deep neural networks for the treatment of sepsis \cite{raghu2017continuous} and   [Rudary et al., 2004] combine RL with constraints for reminder support \cite{rudary2004adaptive}. The latter also shows several forms of personalization that result from learning from patients with different (scheduling) habits.

The work by [Zhu et al., 2017] is related to ours, in that they too focus on clustering the set of users for personalization purposes and use a form of linear function approximation based batch learning as part of their approach \cite{zhu2017group}. In addition to algorithmic differences in learning but also in clustering, a major difference is that we base our experiments on extensive runs with our novel simulator that allows for users that show complex behaviors, have defined behavioral profiles, and thus show much more realistic behavior. We also employ a more sophisticated distance metric in the form of DTW to find optimal clusters of users and use two types of learning in the form of online and batch learning. Some other work exists (cf. the mentioned papers) but so far, most are limited to a few datasets and relatively simple methods. The work by [Raghu et al., 2017] is already a step to employ more advanced methods based on deep learning \cite{raghu2017continuous}, but many other recent techniques in RL will be applicable for e-Health applications (cf. \cite{li2017deep, el2019end}).

Our work is also related to \emph{multi-task} RL, where the goal is to learn policies for multiple problems simultaneously. Some work model an explicit distribution over problems \cite{wilson2007multi}, or \emph{distill} a general policy which can be made more specific \cite{NIPS2017_7036}. In contrast, we focus on clustering groups of users that are alike and learning separate, more specialized policies. Our work is also related to \emph{transfer learning} \cite{taylor2009transfer} where learned policies can be transferred to other tasks, in our case from group level to subgroup level.

%% file: approach.tex
\section{Methodology}
In general terms, our goal is to learn an \emph{intervention strategy} (i.e. a RL policy) for a group consisting of different \emph{users}. In our setting, which types exist, how their behaviour varies, and how different their responses are to the system's intervention, should be unknown beforehand. In our approach, we utilize existing \emph{model-free} RL algorithms to \emph{experiment} with different intervention strategies to improve user's health states. This approach allows us to omit learning models of the environment that would require large amounts of experiences.

\subsection{User Models and Interventions.}
Let $U$ be the set of \emph{users}. We see each user $u \in U$ as a \emph{control problem} modeled as a \emph{Markov decision process} (MDP) \cite{wiering2012reinforcement} $M_u = \langle S_u$, $I$, $T_u$, $R_u \rangle$, where $S_u$ is a finite set of finite \emph{states} the user $u$ can be in, $I$ is the set of possible interventions (\emph{actions}) for $u$, $T_u$ :: $S_u \times I \times S_u \rightarrow$ [0,1] is a probabilistic \emph{transition function} over the states of $u$, and $R_u :: S_u \times I \rightarrow \mathbb{R}$ is a \emph{reward function} that assigns a reward $r = R_u(s_u, i)$ to each state $s_u \in S_u$ and action $i \in I$.

In our system, the set of interventions $I$ contains a binary action as $\{ \mathtt{yes}, \mathtt{no} \}$, representing at each decision moment whether the system sends an intervention or not. The user's state set $S_u$ consists of the \emph{observable} features of the user state. In general, we cannot observe \emph{all} relevant features of the true underlying user state $s_{\mathtt{true}}$ and $S_u$ is therefore restricted to all measurable aspects, modeled through a set of \emph{basis functions} over a state $s_u \in S_u$. That is, we use the feature vector representation $\vec{\phi}(s_u) = \langle\phi_1(s_u),\phi_2(s_u),\ldots,\phi_n(s_u)\rangle^\top$ of the state $s_u \in S$ of user $u$ as representation. If there is no confusion we will use $s_u$ instead of $\vec{\phi}(s_u)$. In our case studies, we choose features that are realistically observable through sensor information, or inferrable.

The transition function $T_u$, which determines how a user $u\in U$ moves from state $s_u \in S_u$ to $s_u^\prime \in S_u$ due to action $i \in I$, is not accessible from the viewpoint of the reinforcement learner, which is a natural assumption when dealing with real human users. In Section ~\ref{sec:simulator}, we do show how we have implemented it for the artificial users in our simulator. The granularity of modeling $T_u$ can be set based on the case at hand, ranging from seconds to hours, denoted $\Delta t$.

Important to note here is that although the time-scale $\delta t$, in reality, can be fine-grained (e.g. $\delta t$ is one second), for the learning algorithms we model $T_u$ at a coarser granularity $\Delta t$ (e.g. $\Delta t$ is one hour): every time point a user $u$ is in some state $s_u \in S$, the system chooses an intervention $i \in I$, upon which the user enters a new state $s_u^\prime$ and a reward $r$ is obtained. Note that for both the transition function and the reward function it is unknown whether they can be considered \emph{Markov}, and thus whether the user can be controlled as an MDP. Nevertheless, we assume it is close enough such that we can employ standard RL algorithms. With a state that is Markov, we can make predictions of future states using only the current state. Note also that all users share the same state representation, but can differ in $R_u$ and $T_u$. An alternative strategy would be to \emph{learn} the dynamics of $T_u$ and $R_u$ from experience as in \emph{model-based} RL (e.g. see~\cite{suttonbarto2nd}), but here we focus on learning them \emph{implicitly} by clustering users who are similar in their behavior (and thus $T_u$ and $R_u$).

\subsection{Evaluating and Learning Interventions.} The goal is to learn intervention strategies, or \emph{policies}, for all users. For any user $u \in U$, $\pi :: S_u \rightarrow I$ specifies the intervention for user $u$ in state $s_u$. The intervention $i = \pi(s_u)$ will cause user $u$ to transition to a new state $s_u^\prime$ and a reward $r = R_u(s_u,i)$ is obtained, resulting in the \emph{experience} $\langle s_u, i, r, s_u^\prime \rangle$. A sequence of experiences for user $u$ can be compactly represented as $\langle s_u, i, r, s_u^\prime, i^\prime, r^\prime, s_u^{\prime\prime}, i^{\prime\prime}, r^{\prime\prime}, \ldots \rangle$ and is called a \emph{trace} for user $u$. For the sake of simplicity we will drop the user subscript if possible. To compare policies, we look at the \emph{expected reward} they receive in the long run. The value of doing intervention $i \in I$ in state $s$ of policy $\pi$, where $\pi(s)=i$, is:
\begin{equation}
	\label{eq:policyeval}
	Q^\pi(s,i) = E_\pi \large\{ \sum_{k=0}^{\infty} \gamma^k r^{t+k+1} | s^t=s, i^t=i \large\}
\end{equation}
\noindent where $\gamma$ is a \emph{discount factor} weighing rewards in the future, and $s^t$ and $i^t$ are states and actions occurring at some future time $t$. From this $Q$-function it is easy to derive a policy, by taking the \emph{best action} $i \in I$ in each state $s \in S$, i.e. 
\begin{equation}
	\label{eq:argmaxQ}
	\pi^\prime(s) = \arg\max_{i \in I} Q^{\pi}(s,i), \ \forall s \in S
\end{equation}
\noindent We are looking for the \emph{best} policy, which is $Q^*(s,i) = \max_{\pi} Q^\pi(s,i)$ for all $s \in S$ and $i \in I$, and $\forall \pi$ for some restricted policy class.

We employ two \emph{off-policy} techniques to learn $Q$-functions: online, table-based $Q$-learning~\cite{watkins1992q} and batch, feature-based \emph{least squares policy iteration} (LSPI)~\cite{lagoudakis2003least}. Let $U$ be our set of users. For $Q$-learning we store each $Q$-value $Q(s,i)$, for $s\in S$ and $i \in I$ separately, and after each experience $(s,i,r,s^\prime)$ for a user $u \in U$ we update the $Q$-function: 
\begin{equation} 
	Q(s,i) \leftarrow Q(s,i) + \alpha \large[ r + \gamma \max_{i^\prime \in I} Q(s^\prime,i^\prime) - Q(s,i) \large]
\end{equation}
\begin{algorithm}[tbp]
\SetAlgoLined
\textbf{Parameters} 

$\alpha \in (0,1]$ is the learning rate 

$\epsilon > 0$ is exploration probability.

\textbf{Initialize}

Q(s,i) $\forall$ $s \in S$, $s \in S$. For terminal states initialize the value with 0.

 \For{\textbf{each} episode}{
  Initialize $s$\
  
  \For{\textbf{each} step in episode}{
  Choose action $i = \pi(s)$ (e.g. using $\epsliongreedy$) \newline
   Take action $i'$, obtain reward $r$ and next state $s'$ \newline
	$Q(s,i) \leftarrow \newline$ \rightline{ $Q(s,i) + \alpha \large[ r + \gamma \max_{i^\prime \in I} Q(s^\prime,i^\prime) - Q(s,i) \large]$} 
   $s \leftarrow s^\prime$

   Stop loop if $s$ is terminal
   }
 }
 \Return{$Q$}
\caption{Q-Learning - An off-policy Temporal-Difference RL algorithm \cite{watkins1992q} }
\label{alg:1}
\end{algorithm}

\noindent where $\alpha$ is the \emph{learning rate}. Note that for all users $U$ together one $Q$-function is learned. Algorithm \ref{alg:1} depicts Q-learning. In addition, we use variants of \emph{experience replay}~\cite{lin1992self} which amounts to performing additional updates by "replaying" experienced traces backward to propagate rewards quicker. In our setting, we sample the experience pairs in chronological order instead of random. Using disjoint experience pairs would have been the better alternative if the set of traces we learn from was larger.

\begin{figure*}
    \centering
    \subfloat{{\includegraphics[scale=0.2,trim=1cm 2cm 1cm 9cm]{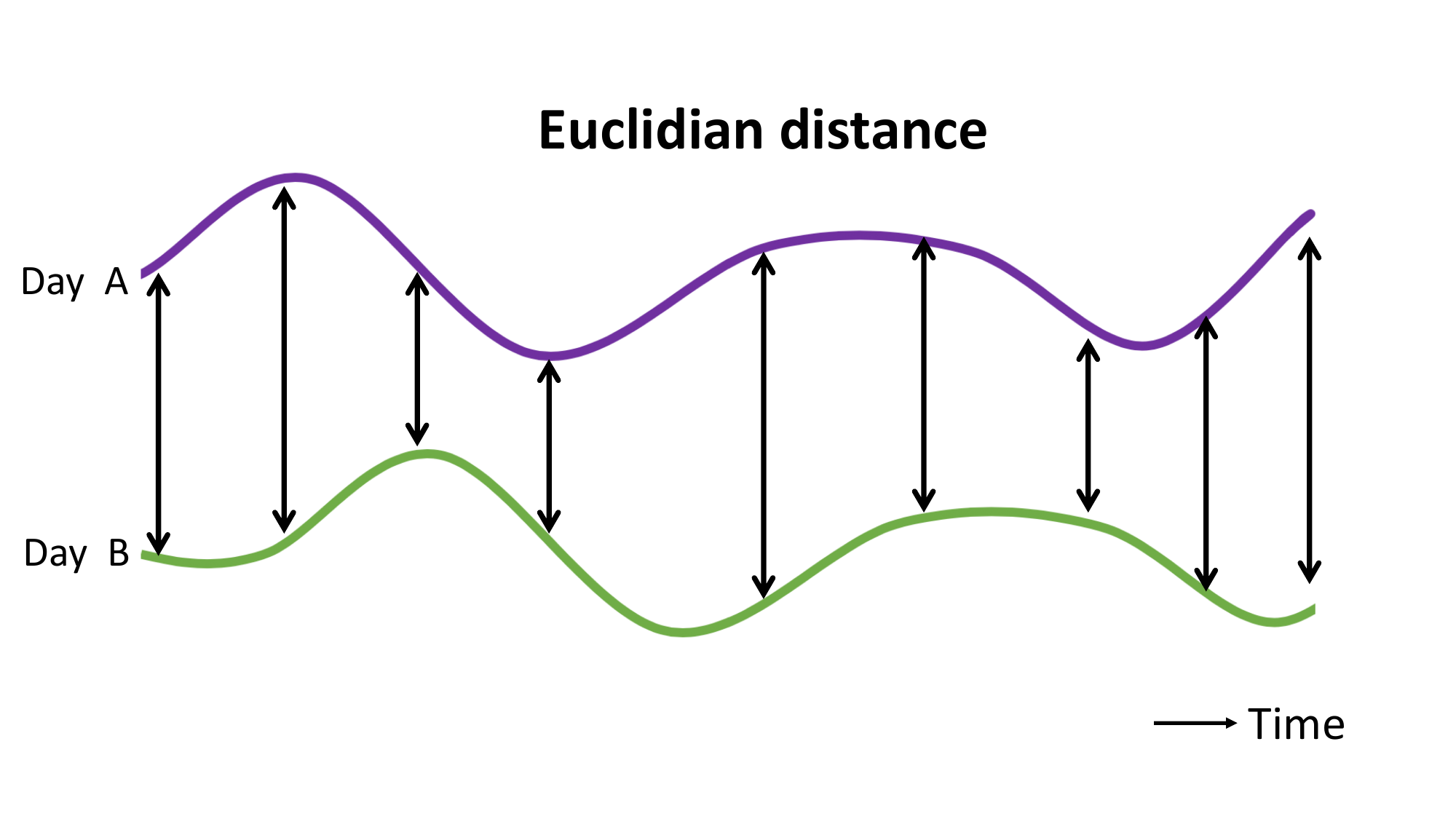}}}  
    \subfloat{{\includegraphics[scale=0.2,trim=1cm 2cm 1cm 9cm]{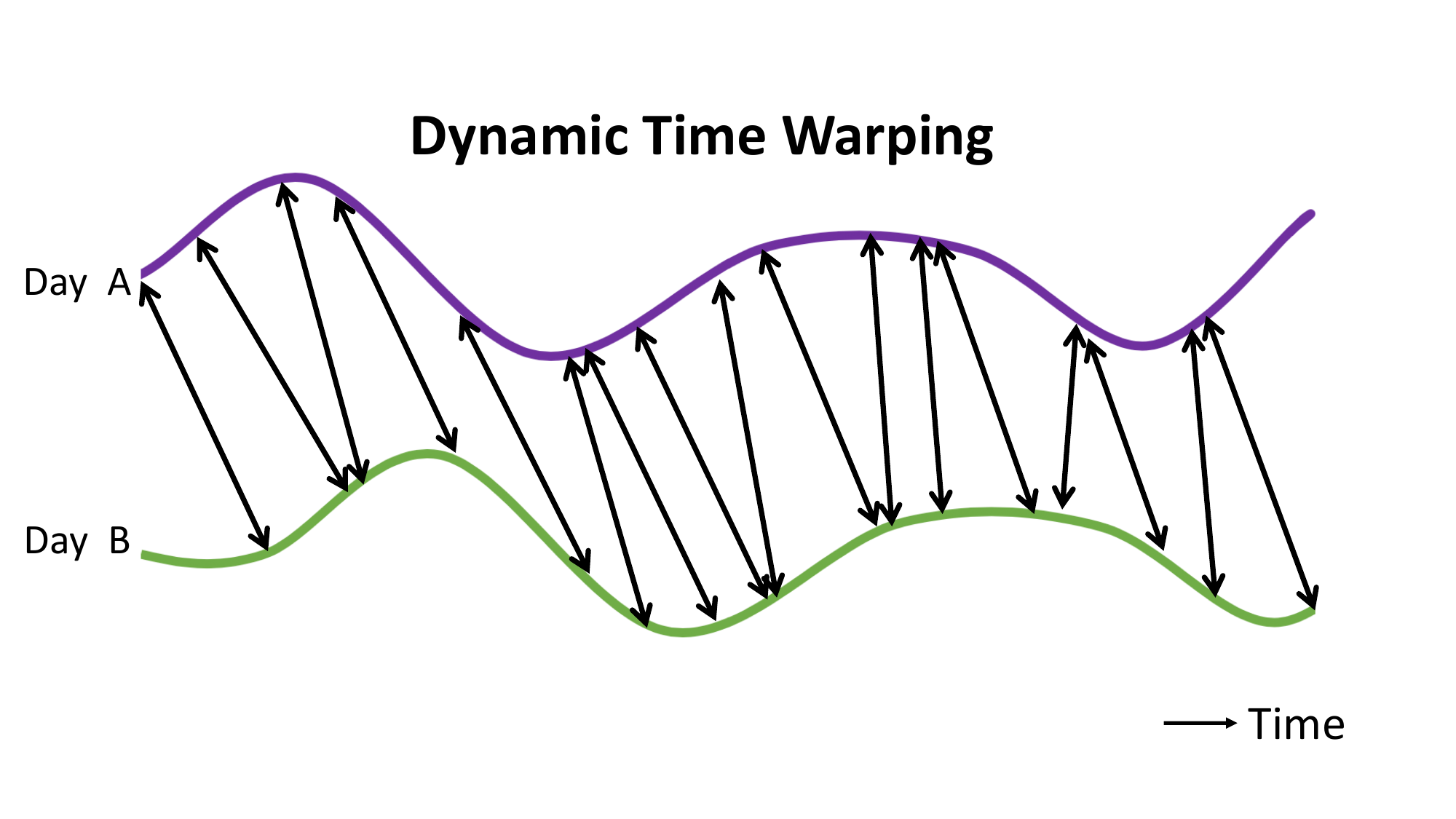}}}
    \caption[abc]{An illustration of the difference between DTW and Euclidean distance applied on the same two sequences that are out of phase.}
    \label{figure_dtw}
\end{figure*}

In our second method, LSPI, we employ the basis function representation $\vec{\phi}(s)$ of a state and compute a \emph{linear function approximation} of the $Q$-function, $\hat{Q} = \sum_{j=1}^{k} \phi(s) w_k$, from a batch of experiences $E$. Here, $\vec{w} = \langle w_1,\ldots ,w_k \rangle$ consists of tunable \emph{weights}. LSPI implements an approximate version of standard \emph{policy iteration} (cf.~\cite{suttonbarto2nd}) by alternating a \emph{policy evaluation step} (Eq~\ref{eq:policyeval}) and a \emph{policy improvement step} (Eq~\ref{eq:argmaxQ}). However, due to the linear approximation, the evaluation step can be computed by representing the batch of experiences in matrix form and using them to find an optimal weight vector $\vec{w}$ using algorithms \ref{alg:2} and \ref{alg:3}. Various methods can be employed for this, and in our experiments we build on the implementation by David Schwab~\footnote{\url{https://pypi.python.org/pypi/lspi-python/1.0.1}}.

\begin{algorithm}[tbp]
\SetAlgoLined
\textbf{Parameters} 

$E$ is the set of experiences (s, i, r, $s^\prime$) 

$k$ is the number of basis functions 

$\vec{\phi}$ is the vector of basis functions 

$\gamma$ is the discount factor 

$\mu$ is the stopping criterion. 

\textbf{Initialize} 

$ E \leftarrow E_0$ (e.g. empty set of experiences) 

$\vec{w}\prime \leftarrow \vec{w_0}$, $\vec{w} \leftarrow \vec{w}\prime$, (default $\vec{w_0} \leftarrow 0$)  
\texttt{\\}

\While{$\neg(\lVert\vec{w} - \vec{w}^{\prime}\rVert)< \mu$}{

Update $E$ (optionally add/remove samples, or leave unaltered). 

$\vec{w} \leftarrow \vec{w}^\prime$

$\vec{w}^\prime \leftarrow \text{LSDQ(E, k,} \vec{\phi}, \gamma, \mu)$
}
\Return $w$
\caption{Least-Squares Policy Iteration (LSPI) - An off-policy RL algorithm \cite{lagoudakis2003least}}
\label{alg:2}
\end{algorithm}

\begin{algorithm}[tbp]

\SetAlgoLined
\textbf{Parameters} 

$E$ is the set of experiences (s, i, r, $s^\prime$)

$k$ is the number of basis functions 

$\vec{\phi}$ is the vector of basis functions 

$\gamma$ is the discount factor 

$\pi$ is the learned policy 

\textbf{Initialize}  

$\widetilde{\textbf{A}} \leftarrow \zeromatrix$ (k x k)   

$\widetilde{b} \leftarrow \zerovector$ (k x 1) 
\texttt{\\}

\For{$\textbf{each} (s, i, r, s^\prime) \in E$}{
$\widetilde{\textbf{A}} \leftarrow \widetilde{\textbf{A}} + \vec{\phi}(s,i)\large[\vec{\phi}(s,i) - \gamma \vec{\phi}(s^\prime,\pi(s^\prime)) \large]^T$ \newline
$\widetilde{{b}} \leftarrow \widetilde{{b}} + \vec{\phi}(s,i)r$
}
$\widetilde{{w}}^\pi  \leftarrow$ $\widetilde{\textbf{A}}^{-1}\widetilde{b}$

\Return $\widetilde{{w}}^\pi$

\caption{LSDQ(E, k, $\vec{\phi}, \gamma, \mu)$ \cite{lagoudakis2003least}}
\label{alg:3}

\end{algorithm}

\subsection{Two Learning Phases.} For any given set of users, we define two phases in learning an optimization strategy. In the first phase (\emph{warm-up}) we employ a default policy $\pi_{\mathtt{def}}$ (see the experimental section for details) to generate traces for each user, and use all experiences of all users to compute $Q^{\pi_{\mathtt{def}}}$. By maximization (Eq.~\ref{eq:argmaxQ}) we obtain a better policy $\pi^\prime$ that is used at the start of the second phase (\emph{learning}). During this phase, we iteratively apply the policy to obtain experiences and update our $Q$-function (and policy) using either $Q$-learning or LSPI. In this phase some exploration is used, reducing the amount of exploration $\epsilon$ over time. After the learning phase, we fix the policy and enter the \emph{performance} phase to evaluate the performance of this final policy. Figure ~\ref{3_phases} provides an overview of the 3 phases warm-up, learning, and performance.

\subsection{Cluster-Based Policy Improvement.} So far, we have assumed all users belong to one group. Our main hypothesis is that since users have different (but unknown) transition and reward functions, learning one general policy for all users will not be optimal. To remedy this, we add a clustering step after the warm-up phase. We employ the K-Medoids clustering algorithm using DTW \cite{Berndt:1994:UDT:3000850.3000887} as the distance metric. Earlier work in e-Health settings has shown that K-Medoids provide good results for clustering users based on behavioral traces \cite{zhu2017group,grua2018exploring}. The advantage of using DTW over the default Euclidean distance is that DTW measures the similarity of two users by calculating the optimal match between the traces of these users, which may be out of phase. The traces that are used here contain the states and reward defined as $\langle s_u, r, s_u^\prime, r^\prime, s_u^{\prime\prime}, r^{\prime\prime}, \ldots \rangle$. To find the optimal match several rules have to be met: (1) every data point from the trace of each user has to be matched with at least one data point from the trace of the other user, (2) the first data point from the trace of the first user has to be matched with that of the second user, (3) the last data point from the trace of the first user has to be matched with that of the second user, and (4) the mapping of the data points from the trace of the first user to those of the second user must increase monotonically. We split the traces of users by day and deploy DTW to calculate the optimal match. We demonstrate the difference between the Euclidean distance and DTW in figure \ref{figure_dtw}. For two signals that are out phase, dynamic time warping will be able to match these signals better leading to a better distance measure. 

Let $U$ be the set of users targeted in the warm-up phase and let $\Sigma^U$ be the set of all traces generated. Let $\Sigma^{\prime u{_{i,m}}}$ be the experiences of user $i$ during day $m$ excluding the interventions. The similarity between users $u_1$ and $u_2$ is defined as:

\begin{equation}
	\label{eq:dtw}
	S_{DTW}(u_1,u_2) = \sum_{m=0}^{M} dtw(\Sigma^{u{_{1,m}}},\Sigma^{u{_{2,m}}}).
\end{equation}

\begin{figure*}[]
\includegraphics[width=\textwidth, trim=1cm 0cm 1.5cm 6.5cm]{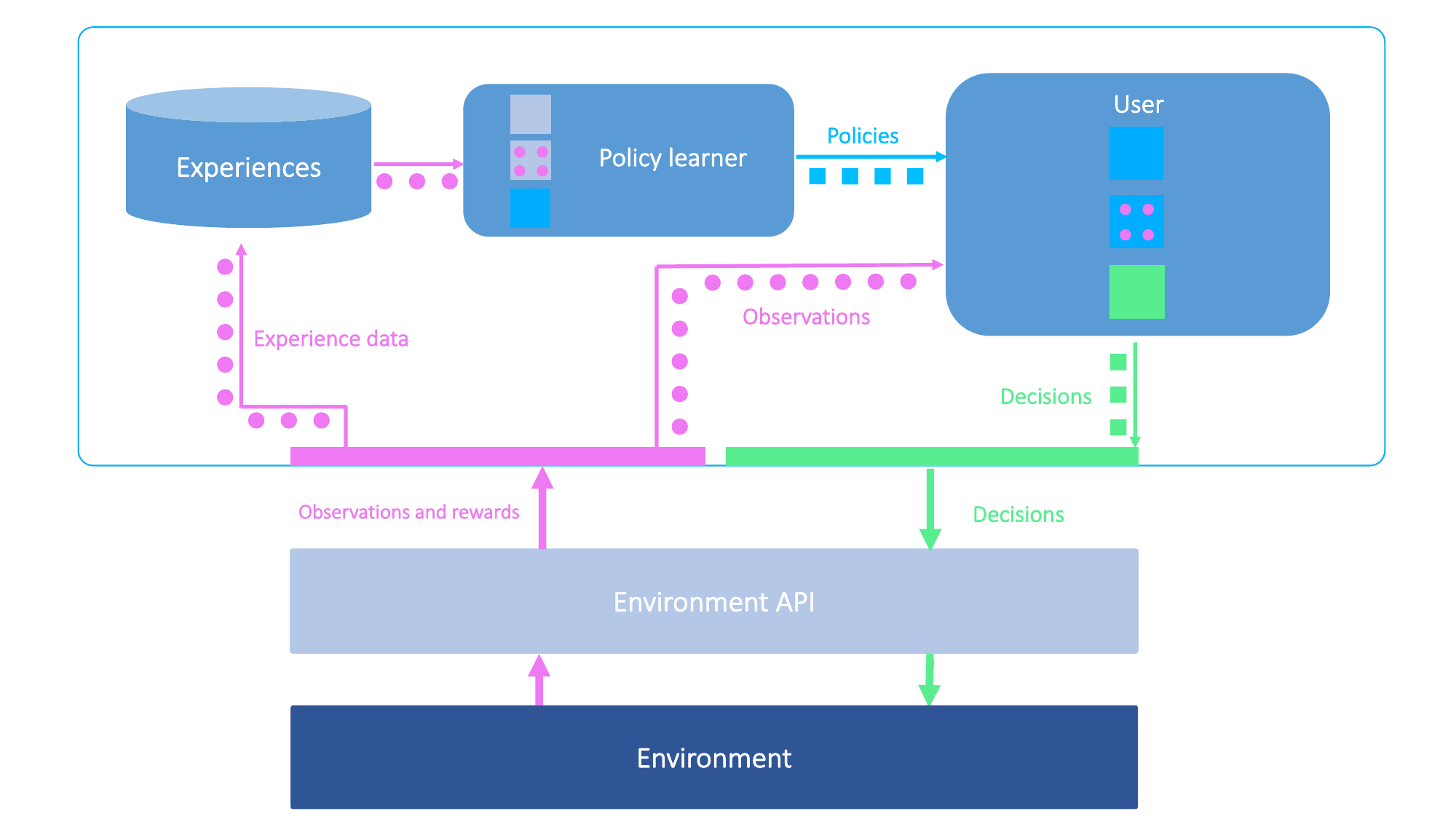}
\centering
\caption{A multi-policy reinforcement learning system for personalized decision-making in e-Health. Given an assignment of users to clusters, policies can be learned across all users, groups of users, or individual users. Algorithms 1, 2, and 3 provide the details for learning the policies.} 
\label{overview_system}
\end{figure*}

Applying the K-medoids algorithm yields a clustering. Let the number of resulting clusters be $k$ and $\Sigma^U_1,\ldots,\Sigma^U_k$ be the partitioning of $\Sigma^U$, and let $U_1, \ldots U_k$ be the partitioning of $U$. Instead of utilizing all experiences of $U$ for one $Q$-function, we now induce a separate $Q$-function $Q_{\Sigma^U_i}$ (and corresponding policy $\pi_{\Sigma^U_i}$) for each user set $U_i$ based on the traces in $\Sigma^U_i$ and continue with learning and performance phases for each subgroup individually. Note that these steps are done in addition to our previous setup, which allows for a comparison between a policy for $U$ and subgroup policies. Figure ~\ref{overview_system} provides an overview of the RL system for personalized intervention in e-Health. For a given setup (i.e. a cluster of users, all users $U$, or per user) an instance of the system described in figure~\ref{overview_system} is created and used to train and update policies.

%% file: evaluation_environments.tex
\section{Evaluation environments}
\label{sec:eval_environments}
Below, we present the evaluation environments for our RL approach. First, we start with a detailed description of the simulator we have developed for this study. Secondly, we describe the HeartSteps benchmark we adopted from literature.

\subsection{An RL multi-agent simulator for e-Health.}
\label{sec:simulator}
For the health setting we focus on in this paper, it is difficult to experiment with different RL strategies and real users, as this requires involving a substantial number of users in a large scale study and gathering too many interaction samples per user. We have therefore decided to build a simulator to experiment with algorithmic settings first \cite{ali_el_hassouni_2020_3826055}. The simulator is created for a realistic setting where users have daily schedules of activities and should be encouraged to conduct certain types of (healthy) activities. In this paper, we rely on data from the US timekeeping research project \cite{hofferth2015american} to define the underlying parameters of the distributions that drive the order of performed activities for the different profiles we define. Below, we discuss the details of the schedules followed by the interventions and the possibility to define rewards.

\subsubsection{Schedules.}
We assume that we have $n$ users in our simulator: $\{u_1,\dots,u_n\}$, originating from the set $U$ as defined before. Each of these users can conduct one of $m$ activities at each time point ($\{\varphi_1,\dots,\varphi_m\}$). Time points in our simulator have a discrete step size $\delta t$. Let $\Phi$ denote the possible values of the activity. Example activities are working, sleeping, working out, and eating breakfast. Each user has a unique activity $a \in A$ that is being conducted at a time point $t$ ($activity: A \times T \rightarrow \Phi$). Note that this activity can also be \emph{none}. For each user, a \emph{template} schedule can be specified, which expresses for each activity $\varphi_i$:

\noindent i) an early and late start time
($early\_start(\varphi_i)$ and $late\_start(\varphi_i)$) with multiple instances per day possible,\\
\noindent ii) a minimum and maximum duration of the activity,
($min\_duration(\varphi_i)$ and $max\_duration(\varphi_i)$)\\
\noindent iii) a standard deviation of the duration of the activity(\emph{$sd\_duration(\varphi_i)$}),\\
\noindent iv) a probability per day of performing the activity ($p(\varphi_i, day)$),\\
\noindent v) priorities of other activities over this activity.

Using these template schedules, a complete schedule is derived which instantiates 
activities at each time point, on a per-day basis, following Algorithm~\ref{alg:simplanning}.

\texttt{\\}
\texttt{\\}

{\scriptsize
\RestyleAlgo{algoruled} 
\LinesNumbered
\begin{algorithm}[tbp]
\SetAlgoLined
\algsetup{linenosize=\tiny}
 day = current\_day\ \newline
 \For{\textbf{each} activity $\varphi_i$}{
 $t_{start}(\varphi_i) = rand(early\_start(\varphi_i), late\_start(\varphi_i))$\
 $d(\varphi_i) = Normal(rand(min\_duration(\varphi_i), $\newline  $ max\_duration(\varphi_i)), sd\_duration)$ \newline
 $p(\varphi_i) = p(\varphi_i, day)$\
 }
 t = \emph{start of the day}\ \newline
 active = false\ \newline
 activity\_queue = \{\}\ \newline
 current\_activity = none\ \newline
 \While{$t <$ end of the day}{
 	activity\_queue = clean\_up\_queue(activity\_queue)\
    \For{\textbf{each} activity $a \in A$}{
       	\If{t == $t_{start}(\varphi_i)$}{
        	\If{rand $\leq$ p($\varphi_i$, day)}{
            	activity\_queue = activity\_queue $\cup$ $\varphi_i$\
            }
        }
    }
	current\_activity = select\_from\_queue(activity\_queue)\
    \If{$\neg$ (current\_activity == none)}{
    	active = true\;
    }
    $t = t + \delta t$
}
\caption{Planning activities per day}
\label{alg:simplanning}
\end{algorithm}}

\texttt{\\}

The algorithm uses the ranges for start times and durations of activities to generate actual start times and durations. The start times are drawn randomly from the specified range and the durations are drawn from a normal distribution with the specified mean and standard deviation. It then starts to run a schedule and builds up a queue of activities that are relevant for the current time point (i.e. for which the current time is after the start of the activity and before the end of it). In case of multiple activities, the one already being performed is continued, or in case of a higher priority activity, the user switches to that activity. If the queue is empty, the user is not active (or idle) and selects the \emph{none} activity.

\begin{figure*}[]
\includegraphics[width=\textwidth,  trim=1cm 10cm 1cm 6cm]{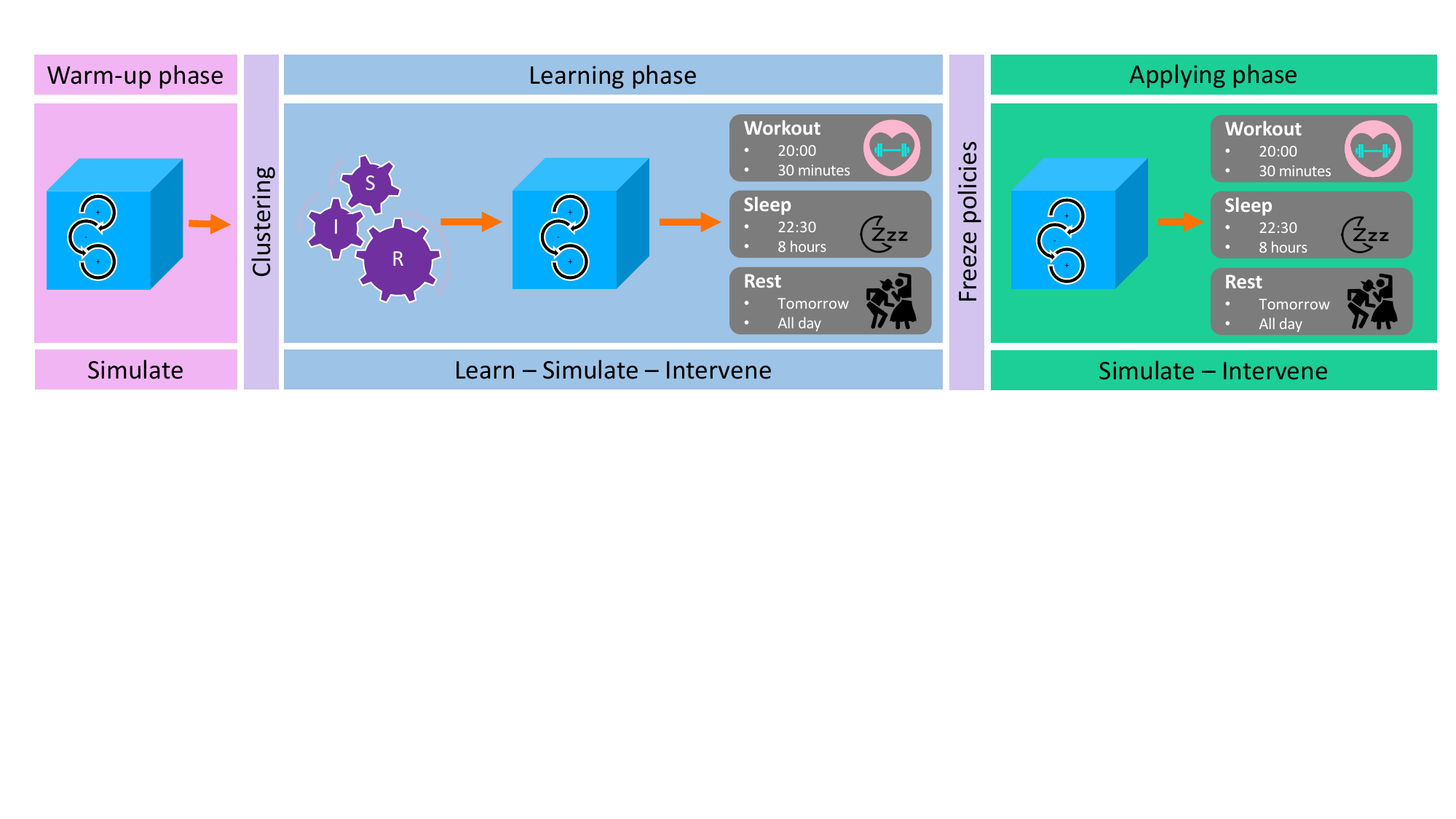}
\centering
\caption{The 3 phases during one simulation run: warm-up, learning, and performance. During the warm-up phase, data is generated following a default policy. Then the clustering step is applied. Using the obtained clustering, policies are learned using the RL system described in figure \ref{overview_system}.} 
\label{3_phases}
\end{figure*}

\subsubsection{Interventions and Rewards.}

Besides performing activities during a day, interventions can also be sent to users. In our system, the set of interventions $I$ contains a binary action as $\{ \mathtt{yes}, \mathtt{no} \}$, representing at each decision moment whether the system sends an intervention or not. An intervention is a message that tells the user to perform a desired activity $\varphi_i$ (we assume there is only one single desired activity for now which is \emph{workout}). To decide upon the acceptance of a message, users have a profile that expresses the conditions under which the users are willing to accept the intervention. These conditions are expressed by the range of time points during which the users are willing to accept an intervention (e.g. a working person might not accept an intervention when at work). If a message is sent at the right time 
and a gap in the schedule is between $t_{plan\_min}$ and $t_{plan\_min} + t_{plan\_duration}$ from the time the message is sent, the activity will be performed. These parameters define a time window in the schedule into which the users will try to fit the desired activity suggested by the intervention. Rewards can be defined based on acceptance of the message (i.e. the activity is considered part of the queue), the commencement of the desired activity, and how long the activity has been performed (e.g. there might be some optimal amount of time spent on the activity). More details on the setting we use for the specific case in this paper are shown in the next section.

\newcounter{mytempeqncnt}

\subsection{HeartSteps.}
\label{heartSteps_generative_model}
We adopt an existing benchmark for e-Health \footnote{\cite{ali_el_hassouni_2020_3824128}: Our implementation of the HeartSteps benchmark} based on the HeartSteps dataset as an additional evaluation of our methods \cite{ali_el_hassouni_2020_3824128}. The dataset was generated during a 42-day long e-Health intervention trial where the goal was to increase the number of steps people take every day by providing interventions in the form of positive messages. These messages, for instance, suggest going for a walk after a long period of sitting \cite{zhu2017group}. We briefly describe this benchmark in this section for the sake of completeness. For a more detailed description of this benchmark and the HeartSteps experiment, we refer to \cite{zhu2017group,lei2014actor,murphy2016batch}. 

\subsubsection{Gaussian generative model.}
Via a micro-randomized trial \cite{lei2014actor,murphy2016batch}, traces of the form $\langle s, i, r, s^\prime, i^\prime, r^\prime, s^{\prime\prime}, i^{\prime\prime}, r^{\prime\prime}, \ldots \rangle$ were collected from users. Using these experiences, a generative model was developed. In this generative model the initial state is drawn from a predefined Gaussian distribution such that $s_{u}(0) \sim$  $N_{p}\large(0,\Sigma \large)$, where $\Sigma$ is a $p$ x $p$ predefined co-variance matrix. In this setting, there are two actions where $1$ indicates a positive intervention and $0$ no intervention. Each of these two interventions is selected from a random policy with probability $0.5$. In our case, there are $3$ numerical states and a numerical reward observable at each time step $t$. 
For each time-step $t \geq 1$, the feature vector representation $\vec{\phi}(s_u)$ for state $s_u$ and the immediate reward are generated using the functions~\ref{eq:generative_model_1}, \ref{eq:generative_model_2}, \ref{eq:generative_model_3}, \ref{eq:generative_model_4}, and \ref{eq:generative_model_5}. Here $ \beta = \large [\beta_i \large]_{i=1}^{14}$ defines the main parameters for each MDP based on the HeartSteps dataset while $\large [ \xi \large ]_{i=1}^{p} \sim \mathcal{N}(0, \sigma_{s}^{2})$ and $\rho_{t} \sim \mathcal{N}(0, \sigma_{r}^{2})$ are the noise distributions for the state and reward models, respectively. 

To generate non-identical experiences for N similar users, N different $\beta$'s need to be created whereby some of these $\beta$'s are closely similar to others forming sets. For user $u$ a $\beta$ is assigned following two steps:

\begin{enumerate}[(a)]
\item Assign user $u$ to basic group $k$ and get the corresponding basic $\beta$ (i.e. $\beta_{k}^{basic}$), \\
\item Make each user $u$ within group $k$ different by adding noise $\delta_u \sim \mathcal{N}(0, \sigma_{\beta}\textbf{I}_{14})$ using $\beta_u = \beta_{k}^{basic}$ + $\delta_{u}$ for $k \in [1,2,\ldots,N_k]$. Here $N_k$ defines the number of users in the $k$-th group and $\textbf{I} \in \mathbb R^{14 x 14}$ an identity matrix.\\
\end{enumerate}

The exact value is chosen for each basic $\beta$ and the parameters of the generative model are discussed in the next section.

\begin{figure*}[]
\normalsize
\setcounter{mytempeqncnt}{\value{equation}}

\begin{equation}
	\label{eq:generative_model_1}
	\leftline{$\phi_{1}(s_u)(t) \leftarrow \beta_{1}\phi_1(s_u)(t-1) + \xi_{t,1},$} 
\end{equation}	

\begin{equation}
	\label{eq:generative_model_2}	
	\leftline{$\phi_{2}(s_u)(t) \leftarrow \beta_2\phi_2(s_u)(t-1) +\beta_3*i_{t-1} + \xi_{t,2},$} \newline
\end{equation}	
	
\begin{equation}
	\label{eq:generative_model_3}	
	\leftline{$\phi_{3}(s_u)(t) \leftarrow \beta_4\phi_3(s_u)(t-1) + \beta_5*\phi_3(s_u)(t-1)*i_{t-1} + \beta_6*i_{t-1} + \xi_{t,3}, $}
\end{equation}	
	
\begin{equation}
	\label{eq:generative_model_4}	
	\leftline{$\phi_{n}(s_u)(t) \leftarrow \beta_7\phi_n(s_u)(t-1) + \xi_{t,n}, n = 4,\ldots,p$} \newline
\end{equation}	

\begin{equation}
\label{eq:generative_model_5}	
\leftline{$r(t) \leftarrow \beta_{14} * \large[\beta_{8} + i_{t} * \large (\beta_{9} + \beta_{10}\phi_{1}(s_u)(t) + \beta_{11}\phi_{2}(s_u)(t) \large) + \beta_{12}\phi_{1}(s_u)(t) - \beta_{13}\phi_{3}(s_u)(t) + \rho_{t} \large]$}
\end{equation}	

\setcounter{equation}{\value{mytempeqncnt}}
\hrulefill
\vspace*{4pt}
\end{figure*}

%% file: experimental_setup.tex
\section{Experimental Setup}
\label{experimental_Setup}

As said, we focus on an e-Health setting whereby learning policies as fast as possible (i.e. based on limited experiences) is essential. The experimental setup is aiming to answer the following questions:
\bigbreak

\emph{\textbf{RQ1}: What are the differences between batch and online learning for our e-Health settings, and how can generalization over state spaces be used to speed up learning?}
\bigbreak

\emph{\textbf{RQ2}: Can a cluster-based RL algorithm learn faster compared to (1) learning per individual user or (2) learning across all users at once?}
\bigbreak

\emph{\textbf{RQ3}: Can we cluster users in a proper way based on traces of their states and rewards?}

\bigbreak

To answer these questions, we evaluate our methodology with two cases, namely: our self developed e-Health simulator and the HeartBeats generative model from the literature.

\subsection{\bf Simulator Setup}
In our simulator setup, we aim to improve the amount of physical activity of users. We include several types of users. More specifically, we employ three \emph{prototypical users}, referred to as the \emph{workaholic}, the \emph{sporter} (an avid athlete), and the \emph{retiree}. The simulator itself runs on a fine-grained time scale ($\delta t$ is one second) while we model $T_u$ at a coarser granularity ($\Delta t$ is one hour). To this end, we rely on the US timekeeping dataset to define the different profiles and their corresponding parameters \cite{hofferth2015american}. 

\subsubsection{Activities.}
We include the following activities: \emph{sleep}, \emph{breakfast}, \emph{lunch}, \emph{dinner}, \emph{work}, \emph{workout}. The specification of the daily schedule for each of our prototypical users is expressed in tables~\ref{tab:params1}, ~\ref{tab:params2}, and ~\ref{tab:params3}. We generate an equal amount of agents for all three types ($n=33$ per type). Each type has its own profile and within each profile we added variability to make sure the agents have some slight differences in preference and behavior.

\begin{table*}[htbp]

  \centering
    \begin{tabular}{c|cccccc}
    Parameter  & \textbf{Sleep} & \textbf{Breakfast} & \textbf{Lunch} & \textbf{Dinner} & \textbf{Work} & \textbf{Workout} \\
    \midrule
    \midrule
    Early start & 22    & 7     & 12    & 18    & 8     & 19.5 \\
    Late start & 23    & 7.5   & 12    & 20    & 9.5   & 20.5 \\
    Min duration & 6     & 0.15  & 0.25  & 0.5   & 9     & 0.5 \\
    Max duration & 7     & 0.25  & 0.5   & 1     & 10.5  & 1 \\
    Priorities & work  & work  & none  & none  & none  & none \\
    Probs (day) & 1,1,1,1,1,1,1 & 1,1,1,1,1,1,1 & 1,1,1,1,1,1,1 & 1,1,1,1,1,1,1 & 1,1,1,1,1,1,0 & 0,0,0,0,0,0,0 \\
    \bottomrule
    \bottomrule
    \end{tabular}%
  \caption{Parameters of the workaholic profile. Start times and durations are in hours.}
  \label{tab:params1}%

\end{table*}%

\begin{table*}[htbp]

  \centering
    \begin{tabular}{c|cccccc}
    Parameter  & \textbf{Sleep} & \textbf{Breakfast} & \textbf{Lunch} & \textbf{Dinner} & \textbf{Work} & \textbf{Workout} \\
    \midrule
    \midrule
    Early start & 21    & 8     & 12    & 19    & 9     & 17 \\
    Late tart & 23    & 9     & 14    & 20.5  & 9.5   & 21 \\
    Min duration & 8     & 0.25  & 0.25  & 0.5   & 8     & 1 \\
    Max duration & 9     & 0.5   & 0.5   & 1     & 8     & 1 \\
    Priorities & work  & work  & none  & none  & none  & none \\
    Probs (day) & 1,1,1,1,1,1,1 & 1,1,1,1,1,1,1 & 1,1,1,1,1,1,1 & 1,1,1,1,1,1,1 & 1,1,1,1,0,0,0 & 0,0,0,0,0,0,0 \\
    \bottomrule
    \bottomrule
    \end{tabular}%
    \caption{Parameters of the sporter profile. Start times and durations are in hours.}
  \label{tab:params2}%

\end{table*}%

\begin{table*}[htbp]

  \centering
    \begin{tabular}{c|cccccc}
    Parameter  & \textbf{Sleep} & \textbf{Breakfast} & \textbf{Lunch} & \textbf{Dinner} & \textbf{Work} & \textbf{Workout} \\
    \midrule
    \midrule
    Early start & 21    & 7     & 12    & 18    & 8     & 15 \\
    Late tart & 23.5  & 10    & 14    & 20    & 9     & 21.5 \\
    Min duration & 8     & 0.5   & 0.25  & 0.5   & 8     & 0.5 \\
    Max duration & 10    & 0.75  & 0.75  & 1     & 8     & 1 \\
    Priorities & work  & work  & none  & none  & none  & none \\
    Probs (day) & 1,1,1,1,1,1,1 & 1,1,1,1,1,1,1 & 1,1,1,1,1,1,1 & 1,1,1,1,1,1,1 & 0,0,0,0,0,0,0 & 0,0,0,0,0,0,0 \\
    \bottomrule
    \bottomrule
    \end{tabular}%
  \caption{Parameters of the retiree profile. Start times and durations are in hours.}
  \label{tab:params3}%

\end{table*}%

\subsubsection{Interventions and Responses.}
The goal of the scenario is to make sure the total workout time meets the guideline for the amount of daily physical activity (30 minutes per day). Messages can be sent to the user to start working out. The acceptance of the message is dependent on the planning horizon of the user and whether it fits into the schedule. The workaholic is a chronic planner, the retiree is a spontaneous planner and the sporter is a mixed planner. The planning horizons in hours of the three \emph{types} are defined as follows: 
(1) \emph{chronic planner} ($t_{plan\_min}$ = 3,  $t_{plan\_duration}$ = 21, $t_{plan\_sd}$ = 0.1), 
(2) \emph{spontaneous planner} ($t_{plan\_min}$ = 0,  $t_{plan\_duration}$ = 6, $t_{plan\_sd}$ = 0.1), and (3) \emph{mixed planner} ($t_{plan\_min}$ = 0,  $t_{plan\_duration}$ = 24, $t_{plan\_sd}$ = 0.1). Here, the standard deviation expresses the variation among the agents spawned for this profile. On top of that, the \emph{workaholic} can only accept interventions when having lunch or being idle while the \emph{retiree} can only accept when idle and the \emph{sporter} always accepts following his acceptance probability. 
The probability of acceptance is set at $0.5$ for the \emph{workaholic}, $0.7$ for the \emph{retiree} and $0.9$ for the \emph{sporter}.

Normally, only one workout per day is performed (and messages can be rejected based on this). However, each of the three types has a probability of working out for a second time in one day. The probability of accepting a second workout intervention is sampled once per agent at the start of the simulation from a normal distribution with parameters $\mu$=$0.05$ and $sd$=$0.05$ for the \emph{workaholic}, $\mu$=$0.05$ and $sd$=$0.05$ for the \emph{retiree} and $\mu$=$0.5$ and $sd$=$0.05$ for the \emph{sporter}. The variations are added to make sure that the behaviors shown by users from the same type are not drawn from the same distributions.  
The sporter has a mean probability of 50\% of working out for a second time during one day, while it is 5\% for both the workaholic and the retiree. 

How long the work out activity will be performed is defined in the profile of the user in Tables~\ref{tab:params1}, \ref{tab:params2}, and \ref{tab:params3}. \emph{Fatigue} plays a role here. Fatigue can build up when working out across multiple days. The value of fatigue is the number of times a user worked out in total during a consecutive number of days where at least one workout per day occurred. A second workout during the same day counts as two workouts in this scenario. When the user skips working out for one day fatigue resets to zero. The maximum value of fatigue is 7. Agents start feeling fatigued after a threshold is reached. This threshold depends on the user. For the retiree, fatigue starts after value 2, for the workaholic after 3, and the sporter after 5. These values are representative of the scenarios we are considering in this setting \cite{rocheleau2004moderators}. The level of fatigue is initialized randomly between 0 and 7 after the start of the simulation. Furthermore, the time that will be spent on a workout is influenced by the level of fatigue. Let $D_{w, u_{i}}(t)$ be the planned duration of the workout at time point $t$ for user $i$ and let $F_{u_{i}}(t)$ be the level of fatigue for user $i$ at time point t. The actual duration $D^\prime_{w, u_{i}}(t)$ for the workout considering the level of fatigue is defined as follows:
\begin{equation}\tag{10}
D^\prime_{w, u_{i}}(t) = \frac{D_{w, u_{i}}(t)}{\sqrt{F_{u_{i}}(t)}}
\end{equation}

\subsection{\bf HeartSteps generative model setup}
In this section we discuss the parameter setup for the HeartSteps generative model. To be able to compare our methods with \cite{zhu2017group}, we adopt the same parameters. For our experiments with the HeartSteps generative model, we select $K = 5$ for the number of groups with each $N_k = 20$ users leading to a total of $100$ users per group and $500$ across all groups. The variance parameters of the Gaussian distributions used to sample noise are 1 for $\sigma_{r}$ and $\sigma_{s}$ and 0.01 for $\sigma_{\beta}$. Furthermore, other parameters have the following values: $p = 3$, and $q = 4$. Finally, the basic $\beta$'s are set in functions~\ref{eq:beta_1}, \ref{eq:beta_2}, \ref{eq:beta_3}, \ref{eq:beta_4}, and \ref{eq:beta_5}. Similar to \cite{zhu2017group}, the number of timesteps $T$ was set to 100 with an evaluation method that averages the long run rewards of all users over a trajectory of length 4000 simulated elements.

\begin{figure*}[]
\normalsize
\setcounter{mytempeqncnt}{\value{equation}}
\setcounter{equation}{10}
\begin{equation}
	\label{eq:beta_1}
	\beta_{1}^{basic} = [0.40, 0.25, 0.35, 0.65, 0.10, 0.50, 0.22, 2.00, 0.15, 0.20, 0.32, 0.10, 0.45, 800] \newline
\end{equation}	

\begin{equation}
	\label{eq:beta_2}	
		\beta_{2}^{basic} = [0.45, 0.35, 0.40, 0.70, 0.15, 0.55, 0.30, 2.20, 0.25, 0.25, 0.40, 0.12, 0.55, 700] \newline
\end{equation}	
	
\begin{equation}
\begin{aligned}
	\label{eq:beta_3}	
		\beta_{3}^{basic} =  [0.35, 0.30, 0.30, 0.60, 0.05, 0.65, 0.28, 2.60, 0.35, 0.45, 0.45, 0.15, 0.50, 650] \newline
\end{aligned}
\end{equation}	
	
\begin{equation}
	\label{eq:beta_4}	
		\beta_{4}^{basic} = [0.55, 0.40, 0.25, 0.55, 0.08, 0.70, 0.26, 3.10, 0.25, 0.35, 0.30, 0.17, 0.60, 500] \newline
\end{equation}	

\begin{equation}
	\label{eq:beta_5}	
		\beta_{5}^{basic} = [0.20, 0.50, 0.20, 0.62, 0.06, 0.52, 0.27, 3.00, 0.15, 0.15, 0.50, 0.16, 0.70, 450] \newline
\end{equation}	

\setcounter{equation}{\value{mytempeqncnt}}
\hrulefill
\end{figure*}

\subsection{\bf Algorithm Setup}\label{algorithm_setup}

In our simulation environment, we instantiate several aspects of our general algorithmic setup from Section 3.

\subsubsection{State.}
As features (i.e. $\vec{\phi}(s_u)$) we use: i) the current time (hours), ii) the current weekday ($0$-$6$), iii) whether the user has already worked out today (binary), iv) fatigue level (numerical), and v) which activities were performed in the last hour (six binary features). All these features are realistically observable through sensor information, or inferable.

\subsubsection{Reward.}
The reward function $R_u$ determines the goal of optimization and consists of three components. If an intervention is sent and the user accepts it, the immediate reward is $+1$ (otherwise $-1$). A second reward component is obtained while the user is exercising, where the exact reward value is scaled relative to the length of the exercise ($+0$ per $\Delta t$) and when the user finishes exercising ($+10$). A third component is related to the fatigue level of the agent at each hour of the day: higher levels result in a small negative reward ($-0.1$ per unit of fatigue per hour) which \emph{shape} the intervention strategy such that it does not overstimulate the user with exercises.

\subsubsection{Default policy.}
The first part of a simulation run is a \emph{warm-up} phase of seven days where interventions are driven by a \emph{default policy} which sends one intervention per day to each user at random between $9$:$00$h and $21$:$00$h. This allows us to perform exploration and to generate traces for clustering.

\subsubsection{Q-learning and LSPI.}
The second part of a simulation run is the \emph{learning phase} that lasts for $100$ days. Immediately after the start of this phase, we update the Q-table using the traces generated during the warm-up phase. In an initial experimentation phase, we tuned several parameters. During the \emph{learning phase} we perform updates to the Q-table once every hour. For Q-learning we use $\gamma=0.95$, and $\epsilon=0.05$ and the learning rate $\alpha$ decreases from an initial $0.2$ with $1\%$ every day. These parameters have been set using grid search for $\gamma$ between $0.85$ and $0.95$ with step size $0.05$, $\epsilon$ between $0$ and $0.05$ with step size $0.05$ and $\alpha$ was fixed at $0.2$ with a $1\%$ decrease rate every day. The total reward was used a the criterion for selecting the parameters. We initialize the Q-values with a random value between $0$ and $1$ if the action of the state-action pair is $0$ otherwise we initialize the Q-values with a random number between $-1$ and $0$, all to encourage exploration. To speed up the learning we use experience replay. We store the last $250$ experiences and use these to update the Q-values. 

For runs with LSPI, we learn policies on the traces generated during the warm-up phase immediately after this phase. The policies get updated at the end of each day by training a new policy on traces from the start of the simulation until that day. For LSPI $\gamma$ was set at $0.95$, $\epsilon$ was selected at $0.01$, the maximum number of iterations was set at $20$ with a threshold of the change in policy weights as a stopping criterion of $0.00001$ and we use a \emph{first win} tie-breaking strategy which returns the first action encountered with that value in case of a tie. Again, parameters have been selected based on a grid search for $\gamma$ between $0.85$ and $0.95$ with step size $0.05$ and for exploration between $0$ and $0.01$ with step size $0.005$. The learning rate and learning rate decay parameters were fixed.

\subsection{\bf Setup of Runs}

We started this section with several research questions. To answer these questions, we run simulations with various configurations. First of all, we vary the usage of the type of RL algorithm: online (Q-learning) and batch learning (LSPI); this enables us to answer \emph{RQ1}. For each type of algorithm, we perform runs where we learn a single policy across all users (\emph{pooled approach}) to a \emph{cluster based} approach and learning a completely individualized policy for each user (\emph{separate approach}). This variation reflects \emph{RQ2}. For our simulation setting, for each algorithm we do two simulation runs for the cluster-based approach; one simulation run using K-Medoids clustering with the DTW distance (\emph{clustering approach}) and a second simulation run using three homogeneous clusters, one for each type of agent (\emph{grouped benchmark approach}). The latter provides us with a (gold standard) benchmark to evaluate the cluster quality (i.e. \emph{RQ3}). Hence, in total, we perform eight runs. For the HeartSteps model, we perform K-Medoids clustering with the DTW distance (\emph{clustering approach}).

%% file: results.tex
\section{Results}

\begin{figure}[]
\includegraphics[width=8cm]{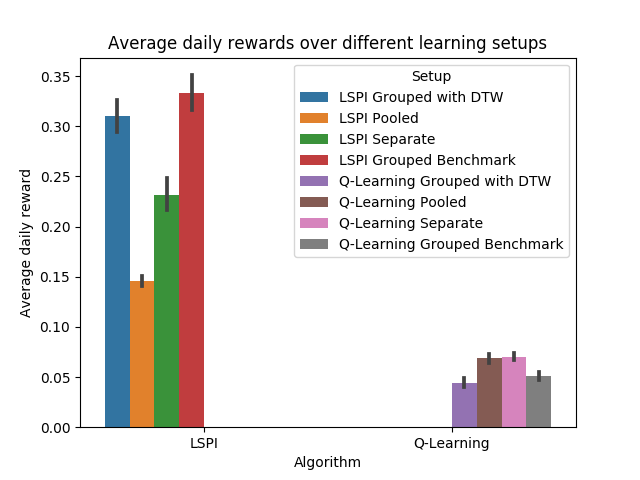}
\centering
\caption{Average rewards over all different setups}
\label{figure1}
\end{figure}

In this section, we present the results related to the three research questions we posed. 

\begin{table*}[]
  \centering

    \begin{tabular}{cccc}
    \toprule
          & Pooled & Grouped  & Separate \\
    \midrule
    \midrule
    HeartSteps & 1291.2 & \textbf{1547.2} & 1435.0 \\
    Batch (LSPI) & 1442.48 & 1446.62 & 1385.12 \\
    Online (Q-learning)  & \textbf{1535.7} & \textbf{1536.71} & \textbf{1536.77} \\
    \bottomrule
    \end{tabular}%
    \caption{The average reward of the three methods pooled, grouped, and separate obtained with the HeartSteps benchmark from \cite{zhu2017group}. The results reported for the HearSteps case in this table are taken from and are based on the implementation of \cite{zhu2017group}. For our implementation see \cite{ali_el_hassouni_2020_3824128}.}
    \label{tab:heartsteps_results}%

\end{table*}%

\subsection{\bf HeartSteps}
Table~\ref{tab:heartsteps_results} shows the results from our runs using our online and batch learning methods on our implementation of the HeartSteps use-case and compares them to the results from \cite{zhu2017group}. Potential discrepancies between our implementation and that from \cite{zhu2017group} are possible. This is due to some details that were missing and the unavailability of publicly accessible implementation of \cite{zhu2017group}. Our results demonstrate that Q-learning (i.e. online learning) outperforms both LSPI (i.e. batch learning) and the benchmark for the pooled case. Also, LSPI outperforms the HeartSteps result in this case. For the grouped approach we see that Q-learning and the HeartSteps achieve comparable results with a slightly better average reward for the HeartSteps benchmark and both outperform the LSPI approach. Finally, for the separate case, we see that Q-learning outperforms LSPI and the HeartSteps with batch learning performing the least of the three. From these results, we can see that the grouped approach always leads to a result closest to the optimal reward. Furthermore, we can see that online learning using Q-learning performs consistently well across all three cases. Furthermore, we observe that our clustering approach finds 4 clusters with a silhouette score of $0.55$. The original work \cite{zhu2017group} does not report on the performance of clustering though it seems. This makes it hard to compare. Also, they fix k (3 and 7) while we look for the optimal k using silhouette score.

\begin{figure}[]
\includegraphics[width=8cm]{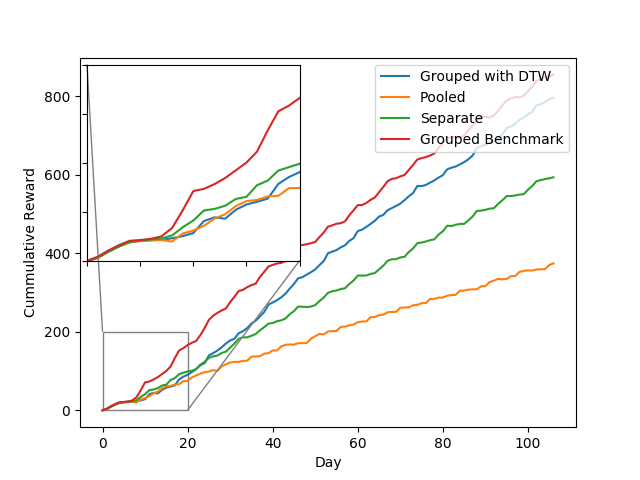}
\centering
\caption{Cumulative reward for batch learning (LSPI) for the different experimental setups obtained with our e-Health simulator.}
\label{CumRewardLSPI}
\end{figure}

\begin{figure}[]
\includegraphics[width=8cm]{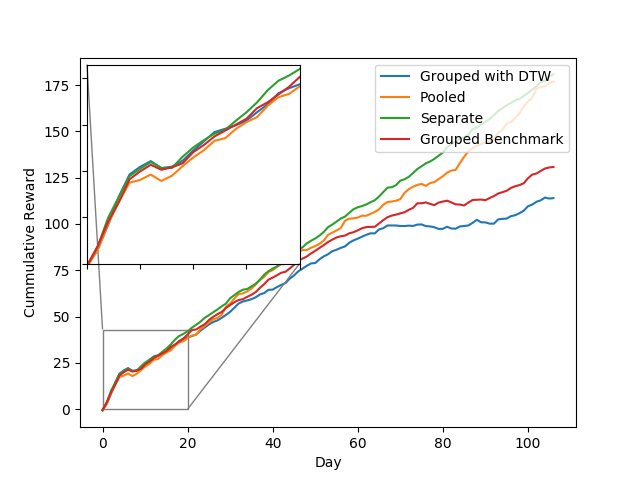}
\centering
\caption{Cumulative reward for online learning (Q-Learning) for the different experimental setups obtained with our e-Health simulator.}
\label{CumRewardQ}
\end{figure}

\subsection{\bf The RL multi-agent simulator for e-Health}
Here, we describe the results obtained with the experiments run with our self-developed simulator for our e-Health setting.

\subsubsection{Batch versus Online Learning.}

Figure~\ref{figure1} reports the results from our simulation runs. Our results demonstrate that LSPI significantly outperforms Q-learning when we compare the average daily reward over the $100$ days during the learning phase. It does so for all four cases (i.e. \emph{separate}, \emph{pooled}, \emph{cluster}, and \emph{grouped benchmark)}. Significance has been tested using a Wilcoxon Signed-Rank test with a significance level of $0.05$. LSPI learned policies that result in average daily rewards between $0.14$ and $0.33$. Q-learning learns policies with average daily rewards of at most $0.065$. The Q-learning experiments show that online (table-based) learning without generalizing over states is not capable of learning reasonable policies in a period of $100$ days (although learning curves show progress, and given excessive amounts of extra time, optimal performance would be reached). LSPI, on the other hand, generalizes over states and utilizes the relatively short amount of interaction much better. This is not a surprise, but it does confirm that generalization -- over the experiences of multiple agents, but also over states -- is needed to obtain reasonable policies in "human-scale" interaction time (and thus answers RQ1).

\subsubsection{Different learning approaches.}
The grouped benchmark approach with LSPI provided us with a policy that outperformed all other policies in this setting. This is, of course, the result of having perfect information about the profiles of the users which allowed us to created perfect clusters. The grouped approach using clustering with DTW was the second-best performing approach and ended very close to the performance of the grouped benchmark approach after learning for 100 days. The separate approach can match the performance of the grouped benchmark approach given enough time to learn. At the same time, the grouped approach outperformed the pooled approach which indicates that clustering helps us learn better policies in a shorter amount of time, by generalizing over the \emph{groups} of agents. We can attribute the slight difference in performance between the clustering approach and the grouped benchmark approach to the fact that the clustering methods we used did not find perfect clusters of the same quality of those of the grouped benchmark approach. However, as shown in Table \ref{tab:clustering}, K-Medoids with DTW finds clusters that are near-optimal. Both the grouped benchmark approach and the separate approach rely on circumstances that are less realistic in the real world. Having more than $100$ days to learn is very difficult and having complete knowledge of the profiles of the users is not realistic. With the clustering-based approach we can speed up the learning time in comparison with the pooled approach to potentially reach better policies.

The policies that were produced by Q-learning show little variation in terms of performance resulting from the different learning approaches. On the contrary, LSPI produces policies learned using the same approaches that are significantly different among each other (Wilcoxon Signed-Rank test, $0.05$ significance). As we can see from Figure.~\ref{figure1}, the policy learned with LSPI using the grouped benchmark approach resulted in the highest average daily reward (Wilcoxon Signed-Rank test, $0.05$ significance). In this case, three clusters were formed each containing precisely the agents of one type. An average daily reward that 1.05 that of the clustering approach and roughly 2.25 times that of the pooled approach was observed. Furthermore, this approach also outperformed the policies learned with a separate approach. Although Q-learning shows little differences across the setups, an interesting observation is that clustering using knowledge about the profiles of the users performs slightly worse in terms of average daily reward than the remaining approach while using Q-learning.

\begin{figure*}[]
\includegraphics[width=\textwidth,  trim=1cm 0cm 1cm 4cm]{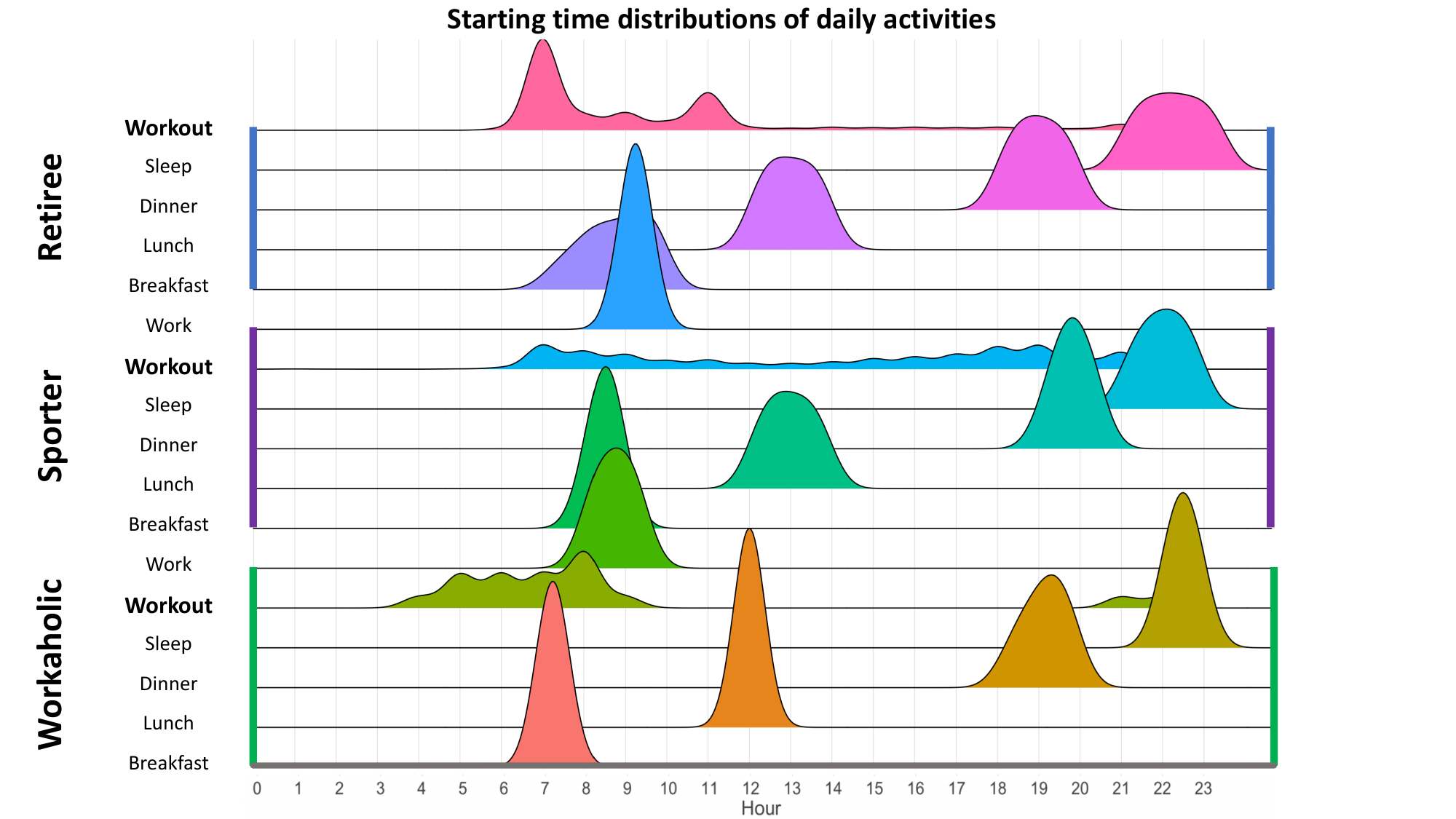}
\centering
\caption{Starting times distributions of activities for the setup: LSPI Grouped with DTW during the last 30 days of the simulation.} 
\label{startingtimedistribution}
\end{figure*}

A different way of measuring performance, by the cumulative average daily reward, is reported in Figures~\ref{CumRewardLSPI} and \ref{CumRewardQ}. These two graphs show the cumulative average daily reward across the different learning setups. For policies learned with LSPI, the grouped benchmark approach provided the highest cumulative reward throughout the simulation in comparison with all other approaches. A small decay was noticeable after $90$ days. The cluster-based approach resulted in a higher cumulative reward throughout the simulation compared to the approaches that learn one policy over all users or rely on learning one policy per user. The pooled approach outperformed the clustering approach during the first 10 days after which the grouped approach was overtaken by the clustering approach.




For the Q-learning case, different behavior was noticeable for the clustering and the pooled approaches. The former gets overtaken by the clustering-based approach after day $20$. The separate benchmark approach provided the lowest cumulative reward throughout the simulation in comparison with all other approaches. The grouped approach is in between these two extremes.

\subsubsection{Clustering.}

Table~\ref{tab:clustering} shows the clustering with the K-Medoids algorithm and the DTW distance metric for the LSPI run. We can clearly see that the clustering is near-optimal for LSPI. Two users of the type \emph{retiree} were confused as the type sporter and one sporter was put together with the workaholics in the same cluster. For the Q-learning case similar patterns were observed.





\begin{table}[H]
  \centering
    \begin{tabular}{c|ccc|ccc}
\cmidrule{2-7}    \multicolumn{1}{r}{} & 1     & 2     & 3     & 1     & 2     & 3 \\ \midrule
    Profile & \multicolumn{3}{c|}{Batch } & \multicolumn{3}{c}{Online } \\
    \midrule
    Workaholic & 0     & 33    & 0     & 33    & 0     & 0 \\
    Sporter & 0     & 1     & 32    & 1     & 1     & 31 \\
    Retiree & 31    & 0     & 2     & 0     & 32    & 1 \\
    \bottomrule
    \bottomrule
    \end{tabular}%
  \label{tab:clustering}%
  \caption{K-medoids clustering performance with DTW. For both setups, near-perfect clustering is found.}
\label{tab:clustering}%
\end{table}%

\subsubsection{In Depth Profile Policy Analysis.}

Figure \ref{startingtimedistribution} reports on the observed starting times of the activities after the simulations have run. We show here data from the last 30 days of the simulation obtained with the setup grouped and algorithm LSPI. We see that the \emph{retiree} mostly works out during morning hours after breakfast, but also after lunch or in the evening. The \emph{sporter} prefers to perform his workout spread over the hours of the day with a higher likelihood during the morning hours and in the evening. Finally, the \emph{workaholic} works out most of the time right after waking up and before having breakfast. There are occasions when the \emph{workaholic} works out before going to bed. These findings indicate that the learned policies accurately learn to send interventions at the right moments and that the users from the different profiles workout more and at the right moments.

\begin{figure*}
    \centering
    \subfloat{{\includegraphics[scale=0.5]{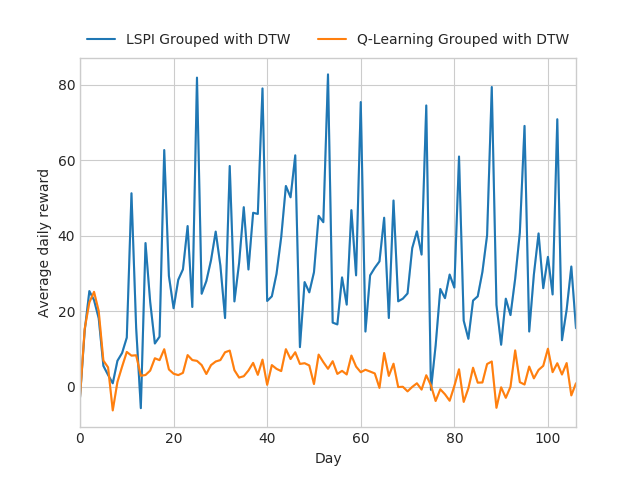}}} 
    \subfloat{{\includegraphics[scale=0.5]{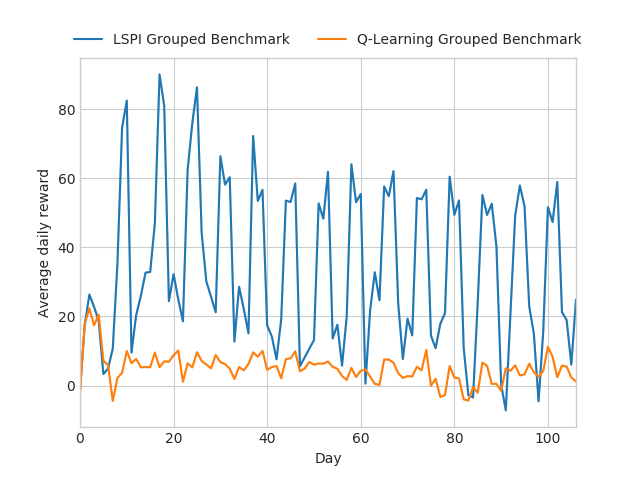}}}
    \qquad
    \subfloat{{\includegraphics[scale=0.5]{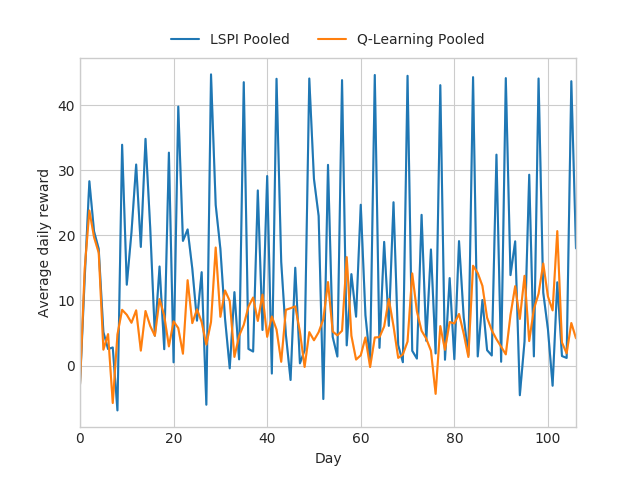}}} 
    \subfloat{{\includegraphics[scale=0.5]{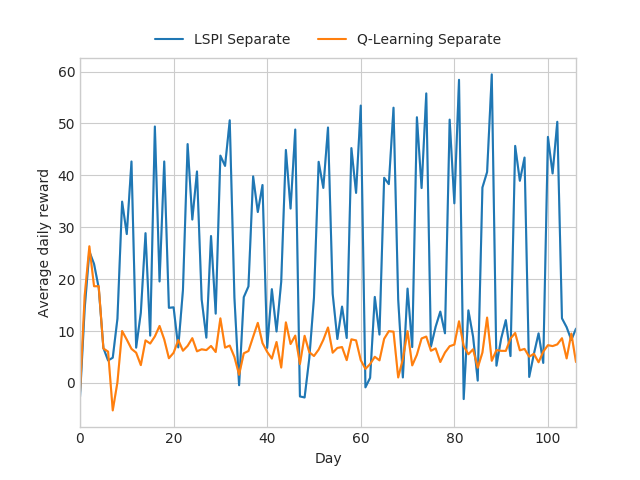}}} 
    \caption[abc]{Average daily reward across all 4 experimental setups (grouped benchmark, grouped, pooled and separate) and the two learning setups (online (LSPI) vs batch (Q-learning)). The steepest increase in average daily reward occurs after 7 days of learning across the different setups.}
    \label{fig:average_daily_reward}
\end{figure*}

Figure \ref{fig:average_daily_reward} reports on the average performance across the different experimental setups and learning algorithms. We see that Q-learning learns slowly, but is consistent over all types of users. LSPI, however, shows great diversity between the different setups in terms of average reward, also which learning setup is most appropriate. We can observe that for LSPI, most of the learning takes place during the first week after which the average reward stabilizes.

Overall, we see that there are three different ways to speed up learning such that learning is feasible in human-scale time: i) generalization over states through basis functions (LSPI) outperforms table-based learning (Q-learning), ii) generalization over traces of several agents (group-based policies) outperforms learning for agents individually (separate learning), and iii) generalization over the \emph{right} agents (cluster-based approaches) outperforms generalization over \emph{all} agents (pooled). All three are needed for interventions in realistic, human domains.

%% file: discussion.tex
\section{Discussion}

In this paper, we have introduced steps towards a cluster-based RL approach for the personalization of e-Health interventions. Such a setting is characterized by limited opportunity to collect experiences from users and where the outcome is focused on optimization of long term health behavior. The presented approach allows for the identification of clusters of users that behave in a similar way and require a similar policy. We have posed various research questions to evaluate the suitability of the approach. Based on the results generated using our novel simulator, for our setting we can say the following. 

\bigbreak

\emph{\textbf{RQ1}: What are the differences between batch and online learning for our e-Health settings, and how can generalization over state spaces be used to speed up learning?}

\bigbreak


RL with batch learning and function approximation outperforms table-based RL using online learning in a significant way, thereby disqualifying the latter when interaction time is short for our e-Health setting using our simulator. For the HeartSteps setting, we observe that online learning outperforms batch learning. Comparing the HeartSteps generative model with our e-Health setting, we can state that our setting allows for more complex behaviours and dynamics of the simulated users with a state-space containing contextual information. Also, our e-Health simulator has a higher level of stochasticity and randomness built-in compared to the HeartSteps setting. 

\bigbreak

\emph{\textbf{RQ2}: Can a cluster-based RL algorithm learn faster compared to (1) learning per individual user or (2) learning across all users at once?}

\bigbreak

In our e-Health setting, cluster-based RL learns a significantly better policy within $100$ days compared to learning per user and learning across all users, provided that a suitable clustering is found. For the HeartSteps setting, the benchmark and batch learning outperform the two settings separate and pooled. However, we find that online learning always performs close to optimal. 

\bigbreak

\emph{\textbf{RQ3}: Can we cluster users in a proper way based on traces of their states and rewards?}

\bigbreak

Learning suitable clusters using the Dynamic Time Warping distance function and K-Medoids clustering based on traces of states and rewards over $7$ days shows to perform very well and find close to optimal clusters for our simulator setup. For the HeartSteps model, 4 clusters with a silhouette score of $0.55$ were found.
\bigbreak

While our simulator exhibits realistic behavior, we plan on moving more and more to a setting where the actual user is in the loop. A logical next step which is to use data collected from actual users to drive the behavior of the agent. We envision to do this by applying machine learning on the data per user and using the resulting model as a behavioral model for that specific user. We already have access to data obtained from a mobile treatment app used by around $250$ depressed patients. In the data, responses to interventions of individual agents are stored as well as socio-demographic and intake questionnaire data and daily ratings of their mental state. Clustering could even be based on the data collected at the start of the intervention. Also, using a state representation that exists of raw sensor has been shown to add to realism of the simulator \cite{el2019end,paper3_end_to_end_RL}. 

%% file: Acknowledgment.tex
\section{Acknowledgment}

This research was supported and co-financed by Mobiquity Inc. We thank Mobiquity Inc for providing the necessary computational resources (AWS) to be able to run the experiments on our simulation environment. 